\documentclass{article}
\usepackage{spconf,amsmath,graphicx}
\usepackage[colorinlistoftodos]{todonotes}
\usepackage{caption}
\usepackage[numbers,sort&compress]{natbib}
\usepackage{subcaption}
\usepackage{bm} 
\usepackage{hyperref}
\usepackage{xcolor}
\usepackage{color}
\usepackage{bm}
\usepackage{url}
\usepackage{subcaption}
\usepackage{booktabs}
\usepackage{multirow}
\usepackage{textcomp}
\usepackage{verbatim}
\usepackage{amssymb}

\usepackage{pifont}       
\usepackage{bbding}       
\usepackage{fontawesome}  
 
\title{A Complementary Global and Local Knowledge Network for Ultrasound denoising with Fine-grained Refinement}

\name{Zhenyu Bu, Kai-Ni Wang, Fuxing Zhao, Shengxiao Li, Guang-Quan Zhou$^*$\thanks{*Corresponding author:guangquan.zhou@seu.edu.cn}}
\address{School of Biological Science and Medical Engineering, Southeast University}
\begin{document}
%
\maketitle
\begin{abstract}
\end{abstract}

Ultrasound imaging serves as an effective and non-invasive diagnostic tool commonly employed in clinical examinations. However, the presence of speckle noise in ultrasound images invariably degrades image quality, impeding the performance of subsequent tasks, such as segmentation and classification. Existing methods for speckle noise reduction frequently induce excessive image smoothing or fail to preserve detailed information adequately. In this paper, we propose a complementary global and local knowledge network for ultrasound denoising with fine-grained refinement. Initially, the proposed architecture employs the L-CSwinTransformer as encoder to capture global information, incorporating CNN as decoder to fuse local features. We expand the resolution of the feature at different stages to extract more global information compared to the original CSwinTransformer. Subsequently, we integrate \textbf{F}ine-grained \textbf{R}efinement \textbf{B}lock (FRB) within the skip-connection stage to further augment features. We validate our model on two public datasets, HC18 and BUSI. Experimental results demonstrate that our model can achieve competitive performance in both quantitative metrics and visual performance. Our code will be available at https://github.com/AAlkaid/USDenoising.

\begin{keywords}
Ultrasound, complementary network, fine-grained refinement, denoising
\end{keywords}
\section{Introduction}
\label{sec:intro}

\begin{figure*}[h]
 \centering
 \includegraphics[width=1\linewidth]{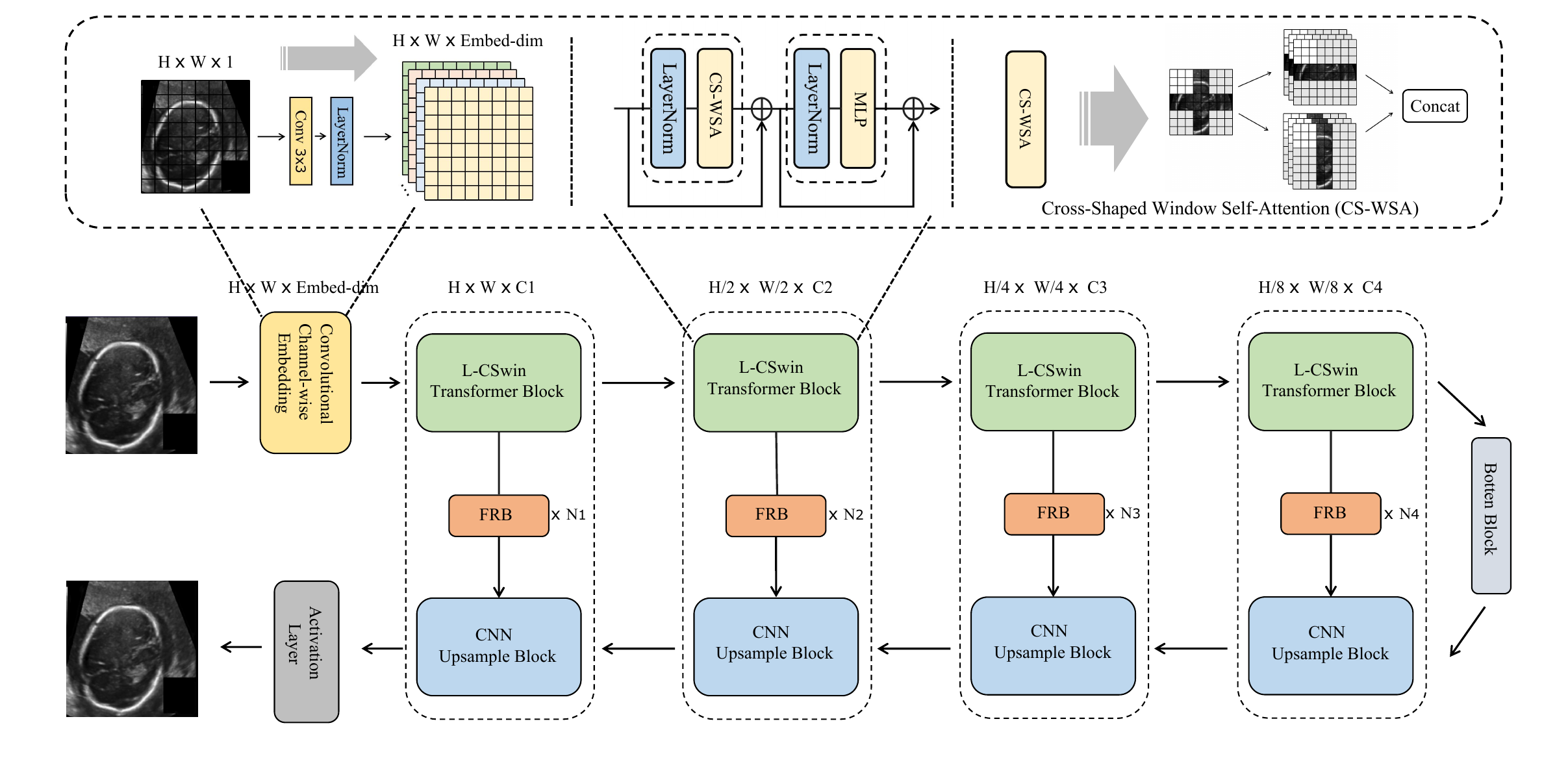}
 \caption{The overall architecture of our proposed network.
  }\label{f2}
\vspace{-2mm}
\end{figure*}

Ultrasound as a cheap and non-invasive examination method has been widely used in various examinations compared with CT and MR \cite{duck2020ultrasound, leighton2007ultrasound}. Although ultrasound examination has huge advantages, ultrasound images are easily affected by speckle noise, which reduces image quality and affects doctors' clinical diagnosis \cite{xie2022improved}. Meanwhile, speckle noise in ultrasound images will greatly affect the performance of downstream tasks, such as segmentation and classification \cite{chen2021lesion, bu2022canet}. Therefore, it is essential to develop an automated and effective ultrasound denoising algorithm.

Over the past several years, a multitude of methods for ultrasound speckle noise reduction have been proposed by researchers. Broadly, these techniques can be categorized into two groups: traditional methods and learning-based methods. 
Dabov et al. proposed a Block Matching and 3D Collaborative Filtering (BM3D)
method for image denoising \cite{dabov2007image}. BM3D mainly uses non-local similarity and sparse representation of images. Michal et al. proposed a K-SVD algorithm, which is an iterative method for sparse representation \cite{aharon2006k}. The goal is to find a "dictionary" matrix such that the input data can be approximated by a few elements in the dictionary. These traditional denoising methods will cause serious blur problems in the image when they come across different noise distributions.

Learning-based methods are widely used in the field of noise reduction due to their good feature extraction capabilities and adaptability to various complex noise situations. There are some CNN-based methods dedicated to solving this problem. Zhang et al. proposed a Denoising Convolutional Neural Network based on convolutional neural networks in 2017 \cite{zhang2017beyond}. The aim of DnCNN is to learn the noise, which can be considered as residual learning. The clean image is obtained by subtraction between the noisy image and the noise. Chen et al. proposed a Residual Encoder-Decoder Convolutional Neural Network (RED-CNN) for low-dose CT denoising \cite{chen2017low}. The main idea is to use an Encoder-Decoder structure to learn a mapping that can convert a low-dose CT image to a high-dose CT image. Nevertheless, the performance of RED-CNN may be suboptimal when confronted with complex tasks. These approaches are ineffective at acquiring global information, resulting in unclear images.

Due to the outstanding ability to capture global features and its excellent performance, the transformer has been widely applied in the field of computer vision, such as ViT, Swintransformer, CSwintransformer \cite{vaswani2017attention, dosovitskiy2020image, liu2021swin, dong2021cswin, wang2022ffcnet, zhou2023tagnet}.
In the field of denoising, there are also many models based on transformer architecture for noise reduction. Liu et al. proposed SwinIR for image restoration \cite{liang2021swinir}, which can be divided into three parts, including the shadow feature extration and deep feature selection, which can better obtain global information through the transformer structure. Wang et al. proposed Uformer, a U-shape pure transformer network for image restoration \cite{wang2022uformer}, which replaces the MLP with Locally-enhanced Feed-Forward (LeFF) to enhance local information. However, existing models cannot fully explore local information, causing the loss of detailed features, which will degrade image quality and impact doctors' clinical judgment.

In order to effectively integrate local and global feature information, as well as enhance detailed feature preservation, we propose a complementary global and local knowledge network for ultrasound denoising with fine-grained refinement. To the best of our knowledge, our proposed method is the first work to combine transformer and CNN for ultrasound image denoising. With the aim of capturing the global information sufficiently, we propose L-CSwintransformer blocks in the encoder, which expand the original size of the feature maps in CSwintransformer blocks from different stages. In the decoder stage, we maintain the classic convolution neural network(CNN) to obtain local information. Furthermore, we introduce fine-grained refinement blocks (FRB) between skip connections to enhance local details of global information to refine the feature map from encoders. Compared with the latest transformer-based method Uformer \cite{wang2022uformer}, our method improves PSNR by 0.4046 and 0.4688 at the noise level 0.25 on the HC18 \cite{van2018automated} and BUSI \cite{al2020dataset} datasets respectively.


\section{Methods}
\label{sec:format}
The proposed complementary network is dedicated to addressing the challenges of over-smoothing and loss of details of ultrasound denoising by combining global and local information, which includes the hybrid architecture and the Fine-grained Refinement Block (FRB). Specifically, we employ the L-CSwinTransformer block to capture global features effectively in the encoder stage. Moreover, we retain the CNN component in the decoder stage in order to achieve local features. Furthermore, we introduce fine-grained refinement blocks (FRB) between the skip connections to further improve the preservation of intricate details. In this section, we will provide a comprehensive description of our proposed method.

\subsection{Architecture overview}
In Figure 1, the proposed model can be divided into three parts: a transformer-based encoder, a CNN-based decoder, and a fine-grained refinement block between the skip connection. The workflow of our proposed network can be described as follows. First, we take a grayscale noisy image as input. The shape of the input is often resized at 224. The original output shapes of the CSwintransformer layers are 56, 28, 14, and 7. In our network, we enlarge the output size of each block to L-CSwintransformer four times. So the output size of each stage is 224, 112, 56, 28. This will help the transformer encoder to better extract the feature information of the entire image. The final output feature is passed through the CNN-based bottleneck and then followed by several CNN upsample blocks. In addition, the output of each stage in the encoder part will also pass through fine-grained refinement blocks to further enhance detail features from the global feature extraction encoder. Finally, the features of each stage after FRB are fused with the upsampled features as complementary features. At the end of the network, we also added a Tanh activation function to constrain the range of the data.
\subsection{L-CSwintransformer encoder}
In order to capture more global features, we proposed a \textbf{L}arge-CSwintransformer as our encoder. CSwintransformer is one of the more popular transformer-based models in 2022. The traditional CSwintransformer uses a large convolution kernel in the embedding stage to reduce the image feature map. In our task, the reduced feature map is not conducive to the extraction of global feature information. Therefore, Our proposed L-CSwintransformer use a small convolution kernel, which can obtain larger feature maps. Therefore, the feature map output by each stage will become larger than the CSwintransformer. Moreover, our proposed L-CSwintransformer uses striped areas to calculate attention and uses striped areas of different widths at different stages of the network to achieve powerful feature modeling capabilities while saving computing resources. The cross-calculation is mainly decomposed into horizontal calculation and vertical calculation, and finally concat together. The utilization of striped areas to compute attention offers a way that mitigates the computational complexity in comparison with direct global attention calculation. Thereby, it establishes a more efficient mode of information interaction.
\subsection{Fine-grained refinement block (FRB)}

\begin{figure}
\centering
\includegraphics[width=0.5\columnwidth, trim={0.1cm 0.1cm 0.1cm 0.1cm}, clip]{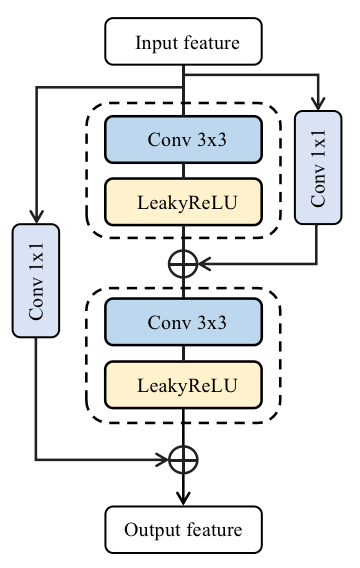}
\caption{Our proposed Fine-grained Refinement Block (FRB).}
\vspace{-0.3cm}
\label{fig:perceiver_tf}
\end{figure}

Our observations revealed that although the global information extraction ability based on the transformer encoder L-CSwintransformer is relatively good. Their ability to enhance details is still lacking. Therefore, we introduced such a Fine-grained Refinement Block (FRB) to further enhance the detailed feature information of the encoder. Since the feature map of the upper layer is larger, we introduce a deeper network, that is, more Fine-grained Refinement Blocks to deepen the feature information. Since the feature map of the lower layer is small and the detailed information is not obvious, we adopt a shallower network structure. As shown in Figure 2, each RFB is composed of Convolution + LeaklyReLU with different convolution kernel sizes and a residual connection composed of 1x1 convolution. In each stage, we use different kernel sizes and different numbers of blocks to enhance feature information. From top to bottom, we reduce the number of blocks from 4 to 1. 
The specific formulaic expression can be expressed as follows: 
\begin{equation}
    \mathcal{O}_{1}= LR(Conv(x)) + {Conv}_{11}(x)
\end{equation}
\begin{equation}
    \mathcal{O}_{2}= LR(Conv({O}_{1})) + {Conv}_{11}(x)
\end{equation}\\
In the above formula, Conv represents convolution kernels of different kernel sizes. Conv11 refers to convolution with kernel size 1, and LR represents LeakyReLU.

\section{Experiments}

\label{sec:pagestyle}
\subsection{Datasets and experimental settings}
Our experiments are conducted on HC18 and BUSI datasets. \textbf{HC18} contains 1334 two-dimensional US images, which are divided into 999 for training and 335 for testing. \textbf{BUSI} contains 780 two-dimensional US images, including 487 benign samples, 210 malignant samples, and 133 normal samples. We divide the BUSI dataset according to the ratio of 7:3.

\begin{figure}
\centering
\includegraphics[width=0.8\columnwidth, trim={0.1cm 0.1cm 0.1cm 0.1cm}, clip]{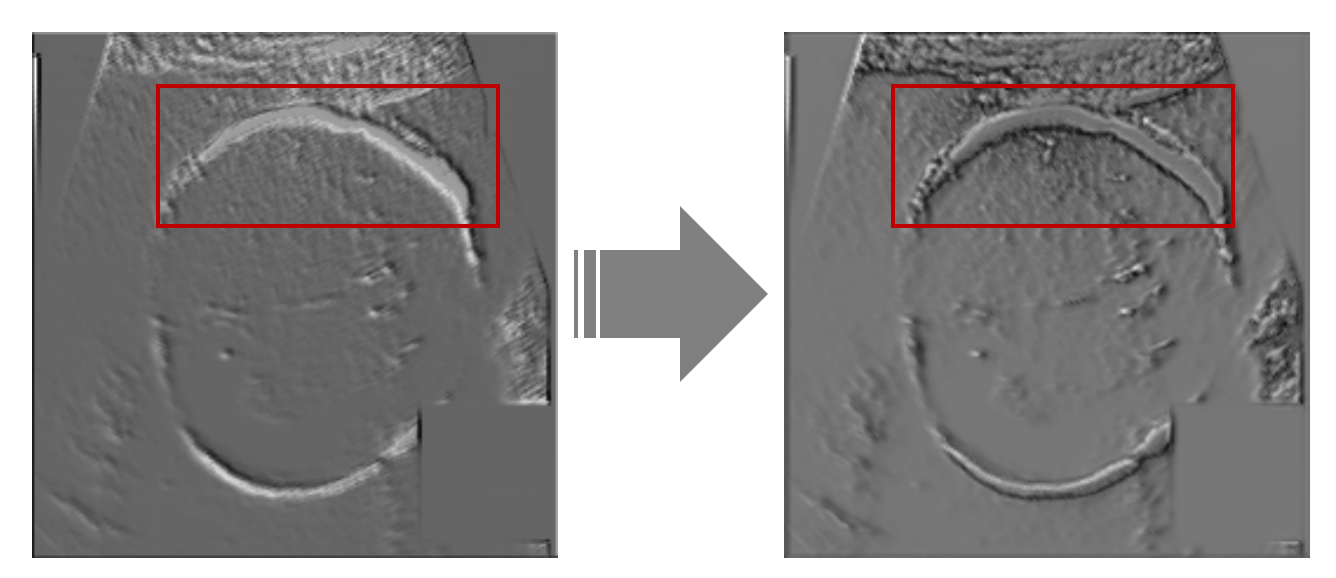}
\caption{Left: Feature map without FRB. Right: Feature map with FRB. The efficacy of the FRB is evident in the comparison between the left and right.}
\vspace{-0.3cm}
\label{fig:perceiver_tf}
\end{figure}

\begin{table*}[t]
\setlength{\tabcolsep}{3.6pt}
\caption{
We present a comparative analysis of denoising results, obtained from the HC18 and BUSI datasets, utilizing diverse methods under varying noise levels. Experimental evaluations were carried out at noise levels of 0.25, 0.35, and 0.45, respectively. The best results across these evaluations are highlighted in bold for clarity and emphasis.
}
\label{table:bss_sdr}\vspace{-0.2cm}
\centering

\begin{tabular}{ @{}l c c c c c c c c c c c@{}} 
\toprule
& & \multicolumn{3}{c}{$\sigma$ = 0.25} & \multicolumn{3}{c}{$\sigma$ = 0.35} & \multicolumn{3}{c}{$\sigma$ = 0.45}\\
 \cmidrule(lr){3-5} \cmidrule(lr){6-8} \cmidrule(lr){9-11} 
Datasets & Methods   & PSNR & SSIM & RMSE & PSNR & SSIM & RMSE & PSNR & SSIM & RMSE\\
\midrule
		
 & BM3D\cite{dabov2007image}  & 31.8299 & 0.8234 & 5.5313 & 30.2134 & 0.8104 & 5.8232 & 30.0021 & 0.8083  & 5.9962 \\ 
& DnCNN\cite{zhang2017beyond}  & 36.6907 & 0.9537 & 3.6983 & 35.3909 & 0.9535 & 4.0566 & 35.3311 & 0.9523 & 4.0533 \\
HC18 & RED-CNN\cite{chen2017low}  & 35.5718 & 0.9557 & 3.7381 & 31.0424 & 0.8968 & 5.1441 & 31.8931 & 0.9367 & 4.5773 \\ 
& SwinIR\cite{liang2021swinir}  & 42.0559 & 0.9899 & 2.0461 & 40.4853 & 0.9877 & \textbf{2.3449} & 38.5431 & 0.9838 & 2.9048 \\ 
& Uformer\cite{wang2022uformer}  & 41.8997 & 0.9878 & 2.0278 & 40.3480 & 0.9873 & 2.4240 & 38.9098 & 0.9840 & 2.7045 \\ 
& Ours  & \textbf{42.3043} & \textbf{0.9904} & \textbf{1.9633} & \textbf{40.6075} & \textbf{0.9884} & 2.3469 & \textbf{38.9764} & \textbf{0.9850} & \textbf{2.6843} \\

\midrule
 & BM3D\cite{dabov2007image}  & 30.4235 & 0.8225 & 7.2388 & 29.9539& 0.8194 & 7.4929 & 28.5893 & 0.8002 & 7.7232 \\ 
& DnCNN\cite{zhang2017beyond}  & 34.2424 & 0.9298 & 5.2323 & 33.0658& 0.9224 & 5.0085 & 32.3621 & 0.9122 & 6.6110 \\ 
BUSI & RED-CNN\cite{chen2017low}  & 32.7425 & 0.9144 & 5.8922 & 31.3452 & 0.9024 & 5.4513 & 30.7340 & 0.8927 & 5.8932 \\ 
& SwinIR\cite{liang2021swinir}  & 36.2255 & 0.9687 & 3.5931 & 35.1947 & 0.9682 & 3.5353 & 34.4113 & 0.9661 & 3.9143 \\ 
& Uformer\cite{wang2022uformer}  & 36.4562 & 0.9699 & 3.5893 & 35.1463 & 0.9670 & 3.6991 & 34.3962 & 0.9666 & 3.9374 \\ 
& Ours  & \textbf{36.9250} & \textbf{0.9704} & \textbf{3.5093} & \textbf{35.2352} & \textbf{0.9688} & \textbf{3.6932} & \textbf{34.6322} & \textbf{0.9675} & \textbf{3.8234} \\ 

\bottomrule
\end{tabular}

\end{table*}

\begin{table}[t]
\caption{Ablation experiments examine the significance of our proposed L-CSwin and FRB. We conducted ablation experiments on the HC18 dataset with 0.25 noise level.}
\label{tab:results_cross_domain}
\centering
\begin{tabular}{p{1.2cm}cccc}
\hline 
\specialrule{0em}{0pt}{1pt}
\multirow{2}{*}{\textbf{L-CSwin}} &
\multirow{2}{*}{\textbf{FRB}} &
\multicolumn{3}{c}{\textbf{}} \\
& &PSNR &SSIM &RMSE\\
\hline 				
\specialrule{0em}{0pt}{1pt}
\XSolidBrush & \XSolidBrush &38.0183  &0.9690 &3.2202 \\
\XSolidBrush  & \Checkmark        &40.5346  &0.9741  &3.0342 \\
\Checkmark & \XSolidBrush         &42.0691  &0.9901 &2.0031 \\
\Checkmark & \Checkmark    &\textbf{42.3043}  &\textbf{0.9904} &\textbf{1.9633} \\
\hline 
\end{tabular}
\vspace{-2mm}
\end{table}

In this study, we use Python and PyTorch. We utilize the L1 as our loss function. The Adam optimization algorithm is employed, with a learning rate initially set at 1e-3. Furthermore, we adopt the ReduceLROnPlateau strategy to perform learning rate decay. Each model is trained for 200 epochs. The experiment is based on the Ubuntu system, with 256GB memory, and two GTX 4090 with 24GB GPU.

In our study, we primarily analyze the experimental results of subjective visual assessment and objective quantitative metrics. The chosen metrics for evaluation include Peak Signal-to-Noise Ratio (PSNR), Structural Similarity Index Measure (SSIM), and Root Mean Square Error (RMSE). 

\subsection{Experimental results}
As displayed in Table 1, Our experiments mainly compared traditional methods, such as BM3D \cite{dabov2007image}, CNN-based methods, such as DnCNN \cite{zhang2017beyond}, RED-CNN \cite{chen2017low}, and transformer-based network architecture SwinIR \cite{liang2021swinir} and Uformer \cite{wang2022uformer}. It is evident that the method we proposed is better than other models with respect to metrics PSNR, SSIM, and RMSE in Table 1. Our approach demonstrates a notable enhancement, increasing the PSNR by 0.4046 on the HC18 dataset and by 0.4688 on the BUSI dataset, both at a noise level of 0.25 in comparison to Uformer.

As illustrated in Figure 4, the visual comparison highlights the notable denoising superiority exhibited by our approach when compared to other methods. When compared with other methodologies, our achieved PSNR surpasses that of the Uformer (d) model by 1.06 on the HC18 dataset and exceeds 1.08 on the BUSI dataset. 

\begin{figure}
\centering
\includegraphics[width=1\columnwidth, trim={0.1cm 0.1cm 0.1cm 0.1cm}, clip]{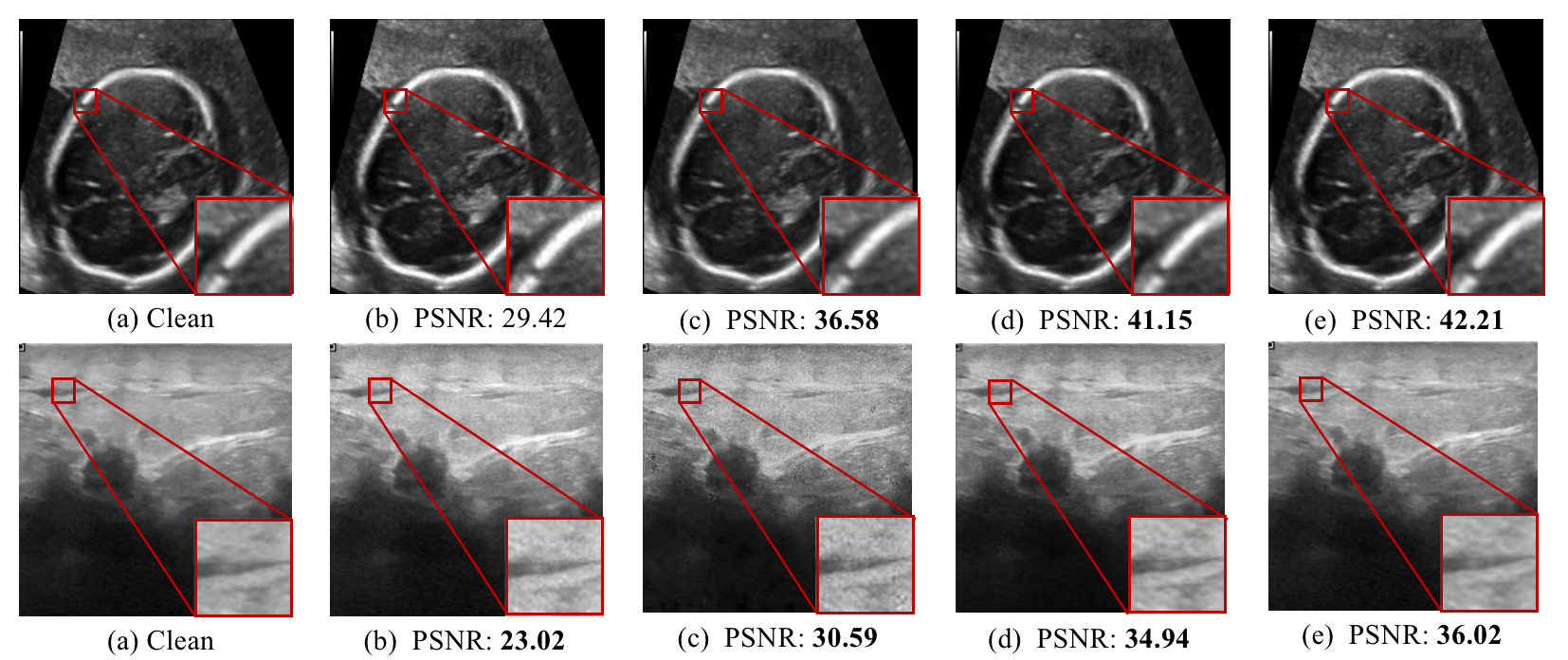}
\caption{Visualization results of HC18 and BUSI.}
\vspace{-0.3cm}
\label{fig:perceiver_tf}
\end{figure}

\subsection{Ablation studies}
Refer to Table 2, we prove the validity of the model by adding or deleting the corresponding L-CSwintransformer and FRB parts. After adding the L-CSwintransformer architecture, the PSNR, SSIM of our hybrid model increased by 4.0608 and 0.0211 respectively. Meanwhile, the RMSE dropped by 1.2171. Hence, the efficacy of our proposed encoder in capturing global information through its transformer-based architecture can be substantiated. After the incorporation of the FRB, there is a noticeable enhancement in all obtained outcomes. Figure 3 proves the effectiveness of our FRB by the visualization. Ultimately, the fusion of the L-CSwintransformer and FRB was undertaken to substantiate the efficacy of our proposed model.

\section{Conclusions}
\label{sec:typestyle}
In this paper, we proposed a complementary global and local knowledge network for ultrasound denoising with fine-grained refinement. Our proposed network is able to extract global feature information comprehensively by the L-CSwintransformer encoder and incorporate it well with the local feature from the CNN upsampling decoder. Furthermore, the Fine-grained Refinement Block (FRB) augments the feature from encoders to obtain clearer results. Experimental results demonstrate that our proposed model outperforms other state-of-the-art methods across various noise levels.

\label{sec:refs}

\bibliographystyle{IEEEbib}

\bibliography{strings,refs}

\begin{thebibliography}{10}

\bibitem{duck2020ultrasound}
Francis~A Duck, Andrew~Charles Baker, and Hazel~C Starritt,
\newblock {\em Ultrasound in medicine},
\newblock CRC Press, 2020.

\bibitem{leighton2007ultrasound}
Timothy~G Leighton,
\newblock ``What is ultrasound?,''
\newblock {\em Progress in biophysics and molecular biology}, vol. 93, no. 1-3, pp. 3--83, 2007.

\bibitem{xie2022improved}
Hui-Wen Xie, Hao Guo, Guang-Quan Zhou, Nghia~Q Nguyen, and Richard~W Prager,
\newblock ``Improved ultrasound image quality with pixel-based beamforming using a wiener-filter and a snr-dependent coherence factor,''
\newblock {\em Ultrasonics}, vol. 119, pp. 106594, 2022.

\bibitem{chen2021lesion}
Kecheng Chen, Kun Long, Yazhou Ren, Jiayu Sun, and Xiaorong Pu,
\newblock ``Lesion-inspired denoising network: Connecting medical image denoising and lesion detection,''
\newblock in {\em Proceedings of the 29th ACM International Conference on Multimedia}, 2021, pp. 3283--3292.

\bibitem{bu2022canet}
Zhenyu Bu, Kaini Wang, and Guangquan Zhou,
\newblock ``Canet: Channel extending and axial attention catching network for multi-structure kidney segmentation,''
\newblock in {\em MICCAI Challenge on Correction of Brainshift with Intra-Operative Ultrasound}, pp. 27--35. Springer, 2022.

\bibitem{dabov2007image}
Kostadin Dabov, Alessandro Foi, Vladimir Katkovnik, and Karen Egiazarian,
\newblock ``Image denoising by sparse 3-d transform-domain collaborative filtering,''
\newblock {\em IEEE Transactions on image processing}, vol. 16, no. 8, pp. 2080--2095, 2007.

\bibitem{aharon2006k}
Michal Aharon, Michael Elad, and Alfred Bruckstein,
\newblock ``K-svd: An algorithm for designing overcomplete dictionaries for sparse representation,''
\newblock {\em IEEE Transactions on signal processing}, vol. 54, no. 11, pp. 4311--4322, 2006.

\bibitem{zhang2017beyond}
Kai Zhang, Wangmeng Zuo, Yunjin Chen, Deyu Meng, and Lei Zhang,
\newblock ``Beyond a gaussian denoiser: Residual learning of deep cnn for image denoising,''
\newblock {\em IEEE transactions on image processing}, vol. 26, no. 7, pp. 3142--3155, 2017.

\bibitem{chen2017low}
Hu~Chen, Yi~Zhang, Mannudeep~K Kalra, Feng Lin, Yang Chen, Peixi Liao, Jiliu Zhou, and Ge~Wang,
\newblock ``Low-dose ct with a residual encoder-decoder convolutional neural network,''
\newblock {\em IEEE transactions on medical imaging}, vol. 36, no. 12, pp. 2524--2535, 2017.

\bibitem{vaswani2017attention}
Ashish Vaswani, Noam Shazeer, Niki Parmar, Jakob Uszkoreit, Llion Jones, Aidan~N Gomez, {\L}ukasz Kaiser, and Illia Polosukhin,
\newblock ``Attention is all you need,''
\newblock {\em Advances in neural information processing systems}, vol. 30, 2017.

\bibitem{dosovitskiy2020image}
Alexey Dosovitskiy, Lucas Beyer, Alexander Kolesnikov, Dirk Weissenborn, Xiaohua Zhai, Thomas Unterthiner, Mostafa Dehghani, Matthias Minderer, Georg Heigold, Sylvain Gelly, et~al.,
\newblock ``An image is worth 16x16 words: Transformers for image recognition at scale,''
\newblock {\em arXiv preprint arXiv:2010.11929}, 2020.

\bibitem{liu2021swin}
Ze~Liu, Yutong Lin, Yue Cao, Han Hu, Yixuan Wei, Zheng Zhang, Stephen Lin, and Baining Guo,
\newblock ``Swin transformer: Hierarchical vision transformer using shifted windows,''
\newblock in {\em Proceedings of the IEEE/CVF international conference on computer vision}, 2021, pp. 10012--10022.

\bibitem{dong2021cswin}
Xiaoyi Dong, Jianmin Bao, Dongdong Chen, Weiming Zhang, Nenghai Yu, Lu~Yuan, Dong Chen, and Baining Guo,
\newblock ``Cswin transformer: A general vision transformer backbone with cross-shaped windows,'' 2021.

\bibitem{wang2022ffcnet}
Kai-Ni Wang, Yuting He, Shuaishuai Zhuang, Juzheng Miao, Xiaopu He, Ping Zhou, Guanyu Yang, Guang-Quan Zhou, and Shuo Li,
\newblock ``Ffcnet: Fourier transform-based frequency learning and complex convolutional network for colon disease classification,''
\newblock in {\em International Conference on Medical Image Computing and Computer-Assisted Intervention}. Springer, 2022, pp. 78--87.

\bibitem{zhou2023tagnet}
Guang-Quan Zhou, Fuxing Zhao, Qing-Han Yang, Kai-Ni Wang, Shengxiao Li, Shoujun Zhou, Jian Lu, and Yang Chen,
\newblock ``Tagnet: A transformer-based axial guided network for bile duct segmentation,''
\newblock {\em Biomedical Signal Processing and Control}, vol. 86, pp. 105244, 2023.

\bibitem{liang2021swinir}
Jingyun Liang, Jiezhang Cao, Guolei Sun, Kai Zhang, Luc Van~Gool, and Radu Timofte,
\newblock ``Swinir: Image restoration using swin transformer,''
\newblock in {\em Proceedings of the IEEE/CVF international conference on computer vision}, 2021, pp. 1833--1844.

\bibitem{wang2022uformer}
Zhendong Wang, Xiaodong Cun, Jianmin Bao, Wengang Zhou, Jianzhuang Liu, and Houqiang Li,
\newblock ``Uformer: A general u-shaped transformer for image restoration,''
\newblock in {\em Proceedings of the IEEE/CVF conference on computer vision and pattern recognition}, 2022, pp. 17683--17693.

\bibitem{van2018automated}
Dhtla Van, DB~Dagmar, K~De, GB~Van, and R~Carlos,
\newblock ``Automated measurement of fetal head circumference using 2d ultrasound images [j],''
\newblock {\em PLoS One}, vol. 13, no. 8, pp. e0200412, 2018.

\bibitem{al2020dataset}
Walid Al-Dhabyani, Mohammed Gomaa, Hussien Khaled, and Aly Fahmy,
\newblock ``Dataset of breast ultrasound images,''
\newblock {\em Data in brief}, vol. 28, pp. 104863, 2020.

\end{thebibliography}

\end{document}